%% file: arxiv.tex
\documentclass[wcp]{jmlr}

\usepackage{longtable}

\usepackage{booktabs}
\usepackage{graphicx}
\usepackage{booktabs}
\usepackage{makecell}
\usepackage{multirow}
\usepackage{algorithm}
\usepackage{algpseudocode}
\usepackage{algorithm2e}
\usepackage{wrapfig} 

\usepackage{ulem}

\usepackage{lineno}

\makeatletter
\let\Ginclude@graphics\@org@Ginclude@graphics 
\makeatother

\jmlrvolume{}

\title[Punching Above Precision]{Punching Above Precision:  Small Quantized Model Distillation with Learnable Regularizer}

\author{\Name{Abdur Rehman} \Email{abdur@opt-ai.kr} \\
\Name{S M A Sharif} \Email{sharif@opt-ai.kr} \\
\Name{Md Abdur Rahaman} \Email{rahaman@opt-ai.kr} \\
\Name{Mohamed Jismy Aashik Rasool} \Email{rasool@opt-ai.kr} \\
\Name{Seongwan Kim} \Email{swan.kim@opt-ai.kr} \\
\Name{Jaeho Lee} \Email{jaeho.lee@opt-ai.kr} \\
\addr Opt-AI, Seoul, South Korea
}

\begin{document}

\maketitle

\begin{abstract}

Quantization-aware training (QAT) combined with knowledge distillation (KD) is a promising strategy for compressing Artificial Intelligence (AI) models for deployment on resource-constrained hardware. However, existing QAT-KD methods often struggle to balance task-specific (TS) and distillation losses due to heterogeneous gradient magnitudes, especially under low-bit quantization. We propose Game of Regularizer (GoR), a novel learnable regularization method that adaptively balances TS and KD objectives using only two trainable parameters for dynamic loss weighting. GoR reduces conflict between supervision signals, improves convergence, and boosts the performance of small quantized models (SQMs). Experiments on image classification, object detection (OD), and large language model (LLM) compression show that GoR consistently outperforms state-of-the-art QAT-KD methods. On low-power edge devices, it delivers faster inference while maintaining full-precision accuracy. We also introduce QAT-EKD-GoR, an ensemble distillation framework that uses multiple heterogeneous teacher models. Under optimal conditions, the proposed EKD-GoR can outperform full-precision models, providing a robust solution for real-world deployment.
\end{abstract}

\section{Introduction}
\begin{figure}[!htb]
  \centering
  \begin{minipage}[t]{0.32\linewidth}
      \centering
      \includegraphics[width=\linewidth, height=3.4cm]{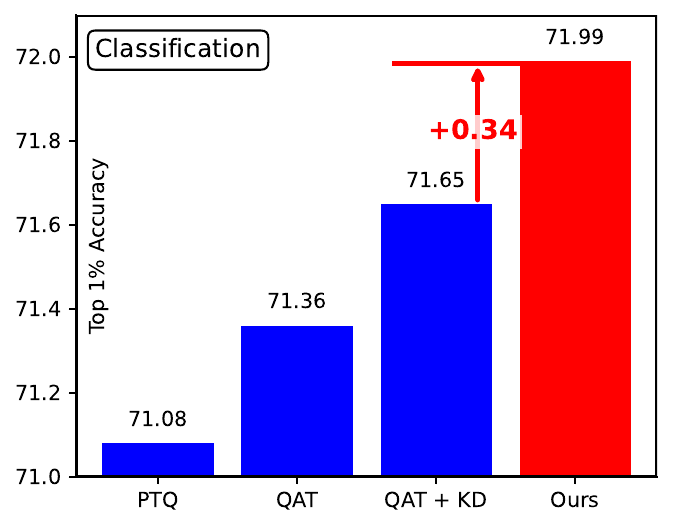}
      \small{(a)} 
  \end{minipage}
  \hfill
  \begin{minipage}[t]{0.32\linewidth}
      \centering
      \includegraphics[width=\linewidth, height=3.45cm]{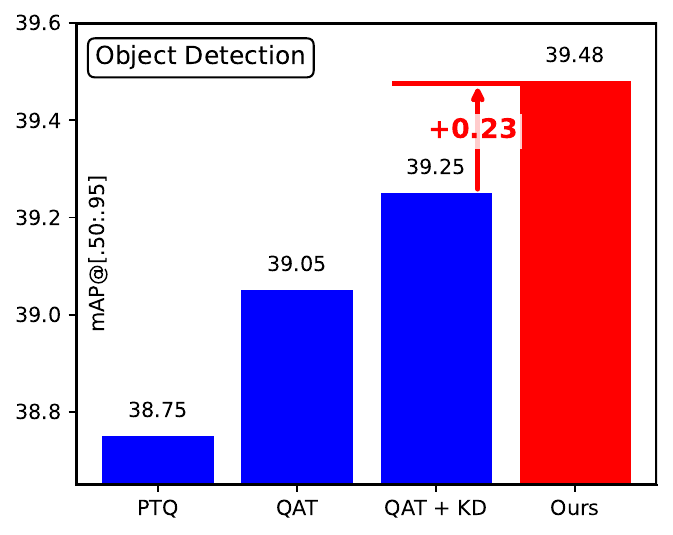}
      \small{(b)}
  \end{minipage}
  \hfill
  \begin{minipage}[t]{0.32\linewidth}
      \centering
      \includegraphics[width=\linewidth, height=3.4cm]{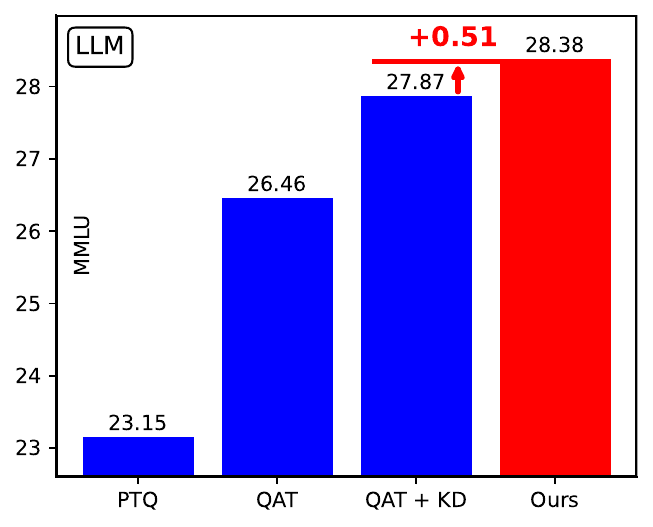}
      \small{(c)}
  \end{minipage}

  \caption{QAT-KD with GoR outperforms traditional PTQ, QAT, and QAT+KD methods. (a) classification (top-1 accuracy), (b) object detection (mAP), and (c) large language model (MMLU)}
  \label{fig:intro}
  \vspace{-0.55cm}
\end{figure}

AI-driven tasks on low-power edge devices are reshaping the next-generation tech ecosystem by enabling critical applications in diverse domains such as healthcare, autonomous systems, and the Internet of Things (IoT)~\citep{rasmussen2021evaluation}. 
However, limited computational and memory resources pose significant challenges for deploying high-performance models in such settings~\citep{a2024learning}. As a result, compact and optimized models are essential for effective deployment in resource-constrained environments. To address this, various model compression techniques such as quantization, pruning, and KD have been explored; yet, achieving an optimal trade-off between efficiency and performance remains an open challenge.

Among popular strategies for model optimization, QAT has emerged as a powerful technique that enables high-precision floating-point parameters to be converted into low-bit integer representations, significantly reducing model complexity and improving inference efficiency~\citep{a2024learning}. However, aggressive quantization often results in substantial accuracy degradation due to unavoidable quantization errors and information loss, limiting its practical effectiveness~\citep{pham2023collaborative}. To address this issue, recent approaches have combined QAT with KD, employing a larger teacher model to transfer robust knowledge representations to a smaller quantized student model, thus mitigating quantization-induced accuracy losses \citep{pham2023collaborative}.

Despite recent advances, QAT-KD methods~\citep{zhu2023quantized, zhao24d} still struggle to balance task-specific (e.g., cross-entropy) and distillation (e.g., KL divergence) losses due to heterogeneous gradient magnitudes and differing optimization dynamics, especially under low-bit quantization. Many recent methods counter these phenomena, relying on extensive hyperparameter tuning to achieve effective convergence~\citep{boo2021stochastic, zhuang2018towards, wang2021collaborative, pham2023collaborative}. Although static or heuristic weighting strategies are commonly used to combine multiple objectives, their lack of adaptability frequently leads to suboptimal learning in SQMs. In contrast, ~\citep{zhao24d} (SQAKD) argue that the TS and KD losses are inherently incompatible. Thus, they propose eliminating TS loss when performing QAT-KD and relying solely on KD guidance instead. Arguably, removing the supervision of TS labels may hinder generalization and degrade performance in real-world scenarios.

The contradictions among existing QAT-KD approaches motivated us to delve deeper into the interaction between TS and KD during QAT. We observe that the conflict between TS and KD is not inherent but rather arises from the absence of an adaptive mechanism to balance them effectively. For instance, completely discarding TS guidance ignores the critical label-driven supervision necessary for semantic alignment and generalization. To address these limitations, we propose GoR, a novel learnable regularization strategy that jointly optimizes both the distillation temperature and loss weighting during training. By adaptively balancing the contributions of TS and KD signals, GoR mitigates quantization-induced performance degradation and fosters better synergy between learning objectives. As shown in Fig.~\ref{fig:intro}, our method consistently outperforms state-of-the-art QAT and QAT-KD baselines, achieving improved generalization and robustness across diverse tasks on low-bit precision (i.e., 8-bit and 4-bit).

 We validated across a variety of compact deep architectures and benchmark datasets. Our results demonstrate its strong generalization ability across diverse AI tasks, such as image classification, OD, and LLM compression. Notably, the proposed QAT-KD-GoR significantly enhances real-world performance on edge hardware, offering up to 242\% faster inference, making it an ideal solution for deployment in resource-constrained environments. Additionally, we explore the potential of Ensemble Knowledge Distillation (EKD), a strategy that combines knowledge from multiple heterogeneous teacher models. Our findings suggest that combining EKD with the GoR mechanism can further boost the performance of SQM in the quantized regime, even outperforming their full-precision counterparts under optimal conditions. In summary, our key contributions are as follows:
\begin{itemize}
    \item We introduce GoR, a novel, learnable regularization technique that dynamically optimizes the balance between TS-loss and KD-loss, significantly improving accuracy and convergence for SQM.

    \item We demonstrate the generalizability of the proposed GoR across diverse AI tasks, including image classification, OD, and LLM. To validate its practical applicability, we evaluate GoR under real-world hardware constraints. Our results show that GoR can enhance the performance of existing methods while introducing negligible additional training parameters.
    
    \item We introduce EKD as a practical solution to address the challenge of lacking homogeneous large-scale teacher models by synthesizing knowledge from heterogeneous large-scale models.
    
\end{itemize}

\section{Related Work}
 \subsection{Quantization-Aware Training}
Quantization methods mainly include Post-Training Quantization (PTQ)~\citep{li2019apot}, which quantizes pre-trained models without retraining but often suffers accuracy loss, and QAT~\citep{krishnamoorthi1806quantizing}, which incorporates quantization during retraining. Early QAT works such as BNN~\citep{courbariaux2016binarized}, XNOR-Net~\citep{rastegari2016xnor}, and DoReFa-Net~\citep{zhou2016dorefa} introduced various scaling techniques, while recent methods employ trainable quantizer parameters to optimize clipping ranges, APoT~\citep{li2019apot}, DSQ~\citep{gong2019differentiable}, and quantization intervals QIL~\citep{jung2019quantization}. Despite improvements, QAT methods vary in effectiveness and lack a unified theory, with MQBench~\citep{li2021mqbench} showing no single method excels universally, especially at low-bit precision.

\subsection{Knowledge Distillation}
 KD methods, introduced by Hinton et al.~\citep{hinton2015distilling}, minimize the KL divergence between teacher and student softmax outputs, alongside cross-entropy loss, and are typically divided into logit-based and feature-based approaches. Logit-based KD, aligning the output predictions of teacher and student networks, is effective for classification and prediction tasks. In contrast, feature-based KD transfers intermediate representations and aligns well with tasks that require richer spatial information, such as detection~\citep{ lightly2025}. Methods like Channel-Wise Distillation (CWD) \citep{shu2021channelwiseknowledgedistillationdense} and Masked Generative Distillation (MGD) \citep{yang2022maskedgenerativedistillation} enhance feature alignment between teacher and student for effective KD on OD. Further research has explored transferring intermediate representations like FSP matrices~\citep{yim2017gift}, attention maps~\citep{zagoruyko2016paying}, internal features~\citep{zhao2023a}, modified distillation loss~\citep {zhou2021rethinking}. A few methods also utilize multiple teachers for effective distillation. For instance, adaptive multi-teacher (AMT) \citep{liu2020adaptive} adaptively weights teacher outputs to generate soft targets, and confidence-aware multi-teacher knowledge distillation (CAMKD) \citep{zhang2022confidence} estimates per-sample teacher reliability using ground truth to guide label fusion.

\subsection{Quantization with Knowledge Distillation}
Recent research uses KD to mitigate accuracy drops in low-precision models. \citep{mishra2018apprentice} introduced Apprentice for ternary-precision and 4-bit networks. QKD \citep{kim2019qkd} divides the process into phases for KD and quantization coordination. SPEQ \citep{boo2021speq} constructs the teacher from the student's parameters with stochastic bit precision. PTG \citep{zhuang2018towards} uses joint training, incremental bit-width reduction, and quantized weight optimization. In QFD \citep{zhu2023quantized}, a quantized teacher model is used for performing distillation on low-bitwidth networks. Collaborative multi-teacher knowledge distillation (CMT) \citep{pham2023cmtkd} encourages mutual learning with multiple quantized teachers. SQAKD \citep{zhao24d} is a self-supervised framework that argues that using only KD loss is sufficient for KD training, achieving state-of-the-art results. Notably, existing KD methods in QAT struggle to balance TS and distillation losses. We address this with a learnable regularizer and extend QAT-KD with QAT-EKD for better performance in low-bit quantization. Beyond discriminative tasks, quantization with distillation has also been investigated for generative models: Q-VDiT \citep{feng2025q} targets video-generation diffusion transformers, while \citet{li2025optimizing} propose cross-timestep error correction for quantized diffusion models. These works highlight the growing breadth of QAT-KD, though our focus is on small quantized models for classification, OD, and LLM compression.

\section{Methodology}

\begin{figure}[!htb]
  \centering

  \begin{minipage}[t]{0.579\linewidth}
      \centering
      \includegraphics[width=\linewidth]{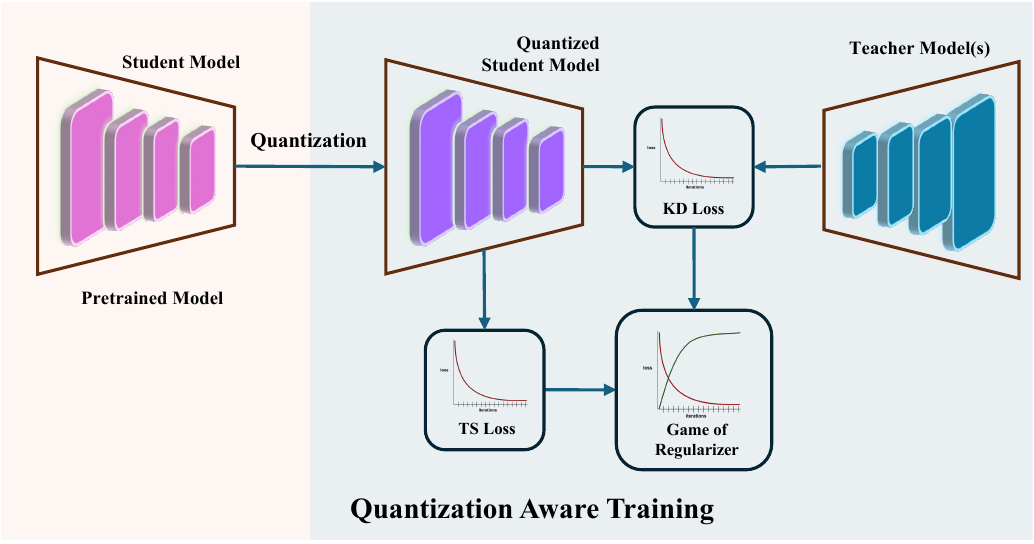}
      \small{(a)} 
  \end{minipage}
  \hfill
  \begin{minipage}[t]{0.413\linewidth}
      \centering
      \includegraphics[width=\linewidth, height=4.65cm]{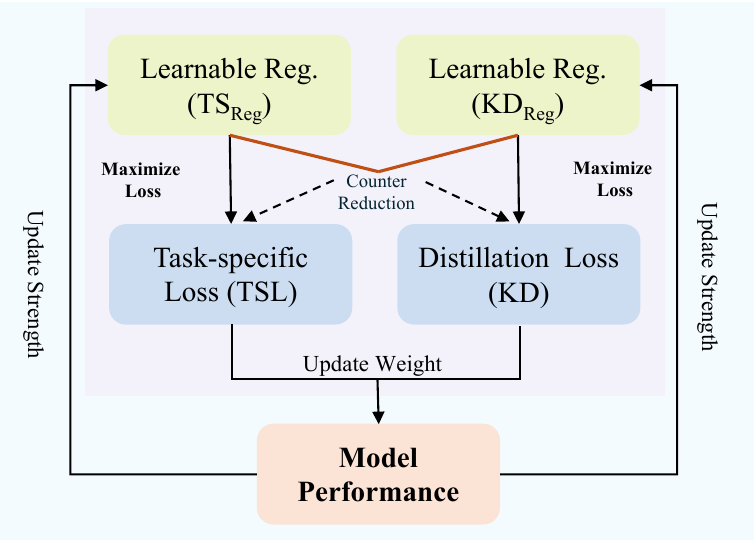}
      \small{(b)}
  \end{minipage}
  \hfill

  \caption{The proposed method learns to balance task-specific and distillation losses during QAT using a learnable regularizer. It employs adaptive components $\text{TS}\text{Reg}$ and $\text{KD}\text{Reg}$ to modulate gradients for stable and effective training. (a) QAT-KD framework with GoR. (b) Details of the proposed GoR.}
  \label{fig:single-cross}
  \vspace{-0.8cm}
\end{figure}

\subsection{Problem Formulation}
KD~\citep{hinton2015distilling} transfers knowledge from a high-capacity teacher network to a lightweight student by matching
their output distributions.
Let \(\mathcal Ds=\{(\mathbf x_i,y_i)\}_{i=1}^{N}\) be a dataset of
inputs \(\mathbf x_i\in\mathbb R^{d}\) and task targets
\(y_i\in\mathcal Y\).
The student is a parametric function
\(f(\mathbf x;\theta)\), and the teacher is
\(f_T(\mathbf x;\theta_T)\) with frozen parameters~\(\theta_T\). We denote by $\mathcal{L}_{\mathrm{task}}$ any differentiable loss appropriate for the downstream task:
\begin{equation}
\mathcal{L}_{\mathrm{task}} = \frac{1}{N}\sum_{i=1}^{N} \ell\!\bigl(f(\mathbf{x}_i;\theta),\,y_i\bigr),
\label{eq:taskloss}
\end{equation}
where $\ell(\cdot,\cdot)$ represents the task-specific loss function.\footnote{Common instantiations include cross-entropy for classification, smooth-$L_1$ for regression, focal loss for OD, etc. The generic formulation accommodates any differentiable task-specific objective.} KD is realized through a loss function that encourages the student model to mimic the teacher's predictions:

\begin{equation}
\mathcal{L}_{\mathrm{KD}} = \mathcal{D}(f(\mathbf{x}; \theta), f_T(\mathbf{x}; \theta_T))
\label{eq:kd}
\end{equation}
where $\mathcal{D}(\cdot, \cdot)$ represents a distance measure between teacher and student outputs.\footnote{For classification tasks, $\mathcal{L}_{\mathrm{KD}}$ is typically implemented as the Kullback--Leibler divergence between temperature-scaled softmax outputs. For regression and other non-classification tasks, appropriate distance measures such as mean squared error or cosine similarity are commonly used.} A conventional KD training step minimizes the weighted sum with a hand-tuned scalar \(\alpha>0\) balancing fidelity to ground-truth targets and mimicry of the teacher.
\begin{equation}
\mathcal{L}
   = (1-\alpha)\mathcal{L}_{\mathrm{task}}
   \;+\;
     \alpha\,\mathcal{L}_{\mathrm{KD}}
\label{eq:joint}
\end{equation}

To deploy on edge hardware, we need to perform QAT, the student’s full-precision parameters \(\theta\) are mapped to low-precision \(\theta_q = \mathrm{Quantize}(\theta)\).
During QAT the forward pass uses \(\theta_q\) (simulating on-device
behaviour) while gradients flow through the straight-through estimator.
Replacing \(\theta\) with \(\theta_q\) in Eq.~\eqref{eq:joint} yields
\begin{equation}
\mathcal L_{\mathrm{QAT}}
   = (1-\alpha)\mathcal L_{\mathrm{task}}
       \bigl(f(\mathbf x;\theta_q),y\bigr)
   \;+\;
     \alpha\,
     \mathcal L_{\mathrm{KD}}
       \bigl(f(\mathbf x;\theta_q),f_T(\mathbf x;\theta_T)\bigr).
\label{eq:qat}
\end{equation}

Quantization introduces non-differentiable noise in the forward pass,
whereas KD tries to smooth the student’s prediction space.  The fixed
coefficient~\(\alpha\) in Eq.~\eqref{eq:qat} is therefore notoriously
fragile: too small, and the student overfits quantization artifacts;
too large and task gradients vanish, causing convergence to stall.  The proposed 
GoR introduced next replaces this
static trade-off with two trainable, mutually balancing scalars that
adapt online to the loss landscape.

\subsection{Game of Regularizers}

\newcommand{\Ltask}{\mathcal{L}_{\mathrm{task}}}
\newcommand{\Lkd}{\mathcal{L}_{\mathrm{KD}}}
\newcommand{\Lgor}{\mathcal{L}_{\mathrm{GoR}}}

\label{subsec:gor}
Motivated by recent observations that fixed loss weighting ~$\alpha$ in Eq.~\eqref{eq:qat}, we introduce the GoR, sketched in Fig.~\ref{fig:single-cross}(b). GoR replaces the static trade-off with two learnable scalars:
\(\alpha_{\mathrm{task}}\) attached to the task loss \(\mathcal{L}_{\mathrm{task}}\) of Eq.~\eqref{eq:taskloss} and \(\alpha_{\mathrm{KD}}\) attached to the distillation loss
\(\Lkd\) of Eq.~\eqref{eq:kd}.  
The two scalars interact antagonistically. Thus, neither can dominate the optimization.
We denote the positive regularizers as
\(
\alpha_{\mathrm{task}},\alpha_{\mathrm{KD}}\in\mathbb R_{>0}.
\)
The GoR objective:
\begin{equation}
\Lgor
   \;=\;
     \frac{\alpha_{\mathrm{task}}}{\alpha_{\mathrm{KD}}}\,
     \Ltask
   \;+\;
     \frac{\alpha_{\mathrm{KD}}}{\alpha_{\mathrm{task}}}\,
     \Lkd.
\label{eq:gor}
\end{equation}

Conventional reweighting strategies such as fixed coefficients, gradient-norm balancing, or uncertainty-based weighting are not directly transferable to the quantized setting. 
In QAT, each loss gradient can be expressed as
\(
\nabla_\theta \mathcal{L}_{i}(\theta_q) 
= g_i(\theta) + \xi_i(\theta),
\)
where $g_i(\theta)$ denotes the clean gradient and $\xi_i(\theta)$ is a parameter-dependent, biased, and heteroskedastic perturbation induced by quantization. 
The scale and distribution of $\xi_i$ differ across the task and distillation losses, distorting both gradient magnitudes and loss statistics. Since existing weighting schemes rely on such statistics, these distortions often cause miscalibration or even invert the intended weighting effect \citep{zhao24d}.

The key insight is that GoR formulation creates a self-regulating system: as one scalar increases to emphasize its loss, it simultaneously reduces the influence of the other loss, preventing either from completely dominating the optimization process. The scalars are optimized along with other model parameters jointly:
\begin{align}
\theta            &\leftarrow \theta - \eta_{\theta}\, \nabla_{\theta}\Lgor \label{eq:theta_update}, \\
\alpha_{\mathrm{task}} 
                  &\leftarrow \alpha_{\mathrm{task}} - \eta_{\alpha}\, \nabla_{\alpha_{\mathrm{task}}}\Lgor \label{eq:alpha_task_update}, \\
\alpha_{\mathrm{KD}} 
                  &\leftarrow \alpha_{\mathrm{KD}} - \eta_{\alpha}\, \nabla_{\alpha_{\mathrm{KD}}}\Lgor \label{eq:alpha_KD_update}.
\end{align}

Here, $\eta_{\theta}$ and $\eta_{\alpha}$ are the learning rates for
parameters and regularizers, respectively, and $\nabla_{x}\Lgor := \partial\Lgor/\partial x$ denotes the gradient of $\Lgor$ with respect to its argument~$x$. After each update we clip the scalars to $[10^{-4},+\infty)$ to guarantee $\alpha_{\mathrm{task}}, \alpha_{\mathrm{KD}} > 0$. Differentiating Eq.\eqref{eq:gor} with respect to each loss term yields:
\begin{align}
\frac{\partial \mathcal{L}_{\text{gor}}}{\partial \mathcal{L}_{\text{task}}} &= \frac{\alpha_{\mathrm{task}}}{\alpha_{\mathrm{KD}}}, &
\frac{\partial \mathcal{L}_{\text{gor}}}{\partial \mathcal{L}_{\text{KD}}} &= \frac{\alpha_{\mathrm{KD}}}{\alpha_{\mathrm{task}}}.
\label{eq:task_kd_derivatives}
\end{align}
While Eq.~\eqref{eq:gor} mathematically depends on the ratio of $\alpha_{\mathrm{task}}$ and $\alpha_{\mathrm{KD}}$, the two-parameter formulation enables unique optimization dynamics through bidirectional competition. Each scalar experiences self-reinforcement and mutual inhibition as shown in ~\eqref{eq:task_kd_derivatives}, creating a game-theoretic equilibrium impossible with single-parameter alternatives like $\mathcal{L} = (1-\beta)\Ltask + \beta\Lkd$. Table~\ref{tab:gorvsnonlearnable} empirically validates this design choice. Analyzing the partial derivatives of~$\Lgor$ with respect to the scalars makes it clearer:
\begin{align}
\frac{\partial\Lgor}{\partial\alpha_{\mathrm{task}}}
 &= \frac{1}{\alpha_{\mathrm{KD}}}\,\mathcal{L}_{\mathrm{task}}
    - \frac{\alpha_{\mathrm{KD}}}{\alpha_{\mathrm{task}}^{2}}\,\mathcal{L}_{\mathrm{KD}}, 
&
\frac{\partial\Lgor}{\partial\alpha_{\mathrm{KD}}}
 &= \frac{1}{\alpha_{\mathrm{task}}}\,\mathcal{L}_{\mathrm{KD}}
    - \frac{\alpha_{\mathrm{task}}}{\alpha_{\mathrm{KD}}^{2}}\,\mathcal{L}_{\mathrm{task}}.
\label{eq:gor_partial_alpha}
\end{align}

Setting these to zero yields the equilibrium
\begin{equation}
\alpha_{\mathrm{task}}^{2}\,
\mathcal{L}_{\mathrm{task}}
\;=\;
\alpha_{\mathrm{KD}}^{2}\,
\mathcal{L}_{\mathrm{KD}}, \label{eq:fixed_point}
\end{equation}
Unlike prior approaches, the coupled formulation in Eq.~\eqref{eq:gor} learns $\alpha_{\mathrm{task}}$ and $\alpha_{\mathrm{KD}}$ jointly with $\theta$, such that the scalars adapt directly to the same biased gradients encountered in QAT-KD. 
The antagonistic balance constraint in Eq.~\eqref{eq:fixed_point} therefore provides stable optimization despite quantization-induced perturbations. Thus, the two losses are power-balanced, allowing each to retain its influence and jointly guide the training process. Equations~\eqref{eq:gor_partial_alpha} reveal the underlying game-theoretic mechanism: each scalar experiences both self-reinforcement (positive terms proportional to its associated loss) and mutual inhibition (negative terms that grow with the competing loss). This creates a natural equilibrium where the scalars dynamically balance to prevent either loss from dominating optimization. Notably, GoR introduces just two trainable scalars, $\mathcal{O}(1)$ parameters, and negligible FLOPs relative to the backbone.

\subsection{Towards QAT–KD-GoR Framework}
\label{subsec:gor_qat_kd}

We propose a GoR-driven QAT–KD framework that dynamically balances task-specific and distillation objectives under quantization constraints. Notably, GoR is modular and can be seamlessly integrated into any existing KD method without requiring architectural modifications or changes to the underlying training pipeline. This makes it broadly applicable across diverse model architectures and tasks, enabling improved performance with minimal overhead. We begin by integrating the GoR weights from Eq.\,\eqref{eq:gor} into the QAT–KD objective of Eq.\,\eqref{eq:qat}, yielding a dynamically balanced loss:
\begin{equation}
  \mathcal{L}_{\mathrm{GoR}}^{\mathrm{QAT}}
    =
      \frac{\alpha_{\mathrm{task}}}{\alpha_{\mathrm{KD}}}
      \,\mathcal{L}_{\mathrm{task}}\bigl(f(\mathbf{x};\theta_q),y\bigr)
    \;+\;
      \frac{\alpha_{\mathrm{KD}}}{\alpha_{\mathrm{task}}}
      \,\mathcal{L}_{\mathrm{KD}}\bigl(
        f(\mathbf{x};\theta_q),\,f_{T}(\mathbf{x};\theta_{T})
      \bigr)
  \label{eq:gor_qat}
\end{equation}
where the learnable scalars \(\alpha_{\mathrm{task}}, \alpha_{\mathrm{KD}} > 0\) dynamically balance task supervision and knowledge distillation during training. To further enhance generalization, we extend knowledge distillation to an ensemble of \(n\) teachers \(\{T_1, \dots, T_n\}\). Let \(\mathbf{z}_{T_i} = f_{T_i}(\mathbf{x}; \theta_{T_i}) \in \mathbb{R}^C\) denote the logits of teacher \(T_i\). We form an ensemble teacher by averaging logits:
\begin{equation}
\mathbf{z}_{\mathrm{ens}} = \frac{1}{n} \sum_{i=1}^{n} \mathbf{z}_{T_i},
\label{eq:ens_logits}
\end{equation}
which is computationally free at inference and empirically competitive with more complex aggregation methods. The student model is then trained to minimize the discrepancy between its predictions and the ensemble teacher's predictions using a distance measure \(\mathcal{D}(\cdot, \cdot)\), adapted for the quantized domain. The generalized EKD loss function can be represented by updating Eq. \eqref{eq:kd} as follows:
\begin{equation}
\mathcal{L}_{\mathrm{EKD}} = \mathcal{D}\left( f(\mathbf{x}; \theta_q), \mathbf{z}_{\mathrm{ens}}\right)
\label{eq:ekd_gen}
\end{equation}
where \(\theta_q = \mathrm{Q}(\theta)\) represents the quantized student parameters. The loss function can be adjusted for various specific tasks by selecting an appropriate distance measure \(\mathcal{D}(\cdot, \cdot)\). Replacing \(\mathcal{L}_{\mathrm{KD}}\) in Eq.\,\eqref{eq:gor_qat} with \(\mathcal{L}_{\mathrm{EKD}}\) from Eq. \eqref{eq:ekd_gen} yields the unified objective:
\begin{equation}
\mathcal{L}_{\mathrm{GoR+EKD}}^{\mathrm{QAT}} =
\frac{\alpha_{\mathrm{task}}}{\alpha_{\mathrm{KD}}}
\, \mathcal{L}_{\mathrm{task}}\bigl(f(\mathbf{x}; \theta_q), y\bigr)
\;+\;
\frac{\alpha_{\mathrm{KD}}}{\alpha_{\mathrm{task}}}
\, \mathcal{L}_{\mathrm{EKD}}
\label{eq:gor_qat_ekd}
\end{equation}
When \(n=1\), \(\mathcal{L}_{\mathrm{EKD}}\) reduces to the conventional KD loss in Eq.\,\eqref{eq:kd}, and Eq.\,\eqref{eq:gor_qat_ekd} recovers Eq.\,\eqref{eq:gor_qat}. Thus, our formulation generalizes classical knowledge distillation to richer teacher ensembles with a learnable weighting scheme.

\begin{algorithm}[!htb]
\DontPrintSemicolon
\caption{Training QAT-KD with Go with optional Ensemble KD}
\label{alg:gor_qat_kd}
\KwIn{%
 dataset $\mathcal{D}$,\;
 pre-trained teacher(s) $\{f_{T_j}\}_{j=1}^n$,\;
 initial student weights $\theta$,\;
 quantization function $\mathrm{Q}(\cdot)$,\;
 temperature $\tau$,\;
 learning rates $\eta_{\theta}, \eta_{\alpha}$}
\tcp{Initialize regularizers}
$\alpha_{\mathrm{task}} \leftarrow 1$, $\alpha_{\mathrm{KD}} \leftarrow 1$ \;
\ForEach{mini-batch $\{(\mathbf{x}_i, y_i)\} \subset \mathcal{D}$}{
    $\theta_q \gets \mathrm{Q}(\theta)$ \tcp*{fake quantization}
    $\mathcal{L}_{\mathrm{task}} \gets \ell\bigl(f(\mathbf{x};\theta_q), y\bigr)$ \;  \tcp*{Compute task loss}
    
    \tcp*{Compute knowledge distillation loss}
    \uIf{$n > 1$ }
        \Indp
        $\mathbf{z}_{T_i} \gets f_{T_i}(\mathbf{x};\theta_{T_i})$ for $i = 1, \ldots, n$ \;
        $\mathbf{z}_{\mathrm{ens}} \gets \frac{1}{n}\sum_{i=1}^{n} \mathbf{z}_{T_i}$ \tcp*{ensemble logits}
        $\mathcal{L}_{\mathrm{KD}} \gets 
        \mathcal{D}(f(\mathbf{x}; \theta_q),\mathbf{z}_{\mathrm{ens}})$  \tcp*{Ensemble KD loss}
        \Indm
    }
    
    \Else{
        \Indp
        $\mathcal{L}_{\mathrm{KD}} \gets \mathcal{D}(f(\mathbf{x}; \theta_q), f_T(\mathbf{x}; \theta_T))$ \tcp*{Standard single-teacher KD}
        \Indm
    }

    $\mathcal{L}_{\mathrm{GoR}} \gets \frac{\alpha_{\mathrm{task}}}{\alpha_{\mathrm{KD}}} \, \mathcal{L}_{\mathrm{task}} + \frac{\alpha_{\mathrm{KD}}}{\alpha_{\mathrm{task}}} \, \mathcal{L}_{\mathrm{KD}}$ \;  \tcp*{Game of Regularizers objective}
    
    $\{\theta_q, \alpha_{\mathrm{task}}, 
    \alpha_{\mathrm{KD}}\} \gets 
    \text{update via equations } \eqref{eq:theta_update}, \eqref{eq:alpha_task_update}, \eqref{eq:alpha_KD_update}$ \;     \tcp*{Update parameters}

    $\alpha_{\mathrm{task}} \gets \max(\alpha_{\mathrm{task}}, 10^{-4})$ \;
    $\alpha_{\mathrm{KD}} \gets \max(\alpha_{\mathrm{KD}}, 10^{-4})$ \;     \tcp*{Clip regularizers}

\KwOut{quantized student $\theta_q^{\star}$ with learned regularizers $\alpha_{\mathrm{task}}^{\star}, \alpha_{\mathrm{KD}}^{\star}$}
\end{algorithm}

\paragraph{Implementation and training.} We implement our framework using NVIDIA's open-source PyTorch-quantization library with fake-quantizer modules to emulate low-bit arithmetic. In our setup, only the student model is quantized during training. Floating-point tensors are scaled by \( s = \frac{x_{\max} - x_{\min}}{2^n - 1} \), rounded, clipped, and dequantized as \( x_q = s \cdot q + x_{\min} \), with gradients approximated via the straight-through estimator. KD is applied between the quantized student and the full-precision teacher (please refer to Fig. \ref{fig:single-cross} (a)), using the KL divergence on temperature-scaled logits and optionally feature-based losses, depending on the distillation method. This allows the student to learn under realistic quantization noise while benefiting from both high-level and intermediate supervision. Algorithm \ref{alg:gor_qat_kd} illustrates the training details of the proposed QAT-KD-GoR as a framework.

\section{Experiments}
\subsection{Comparison with State-of-the-art Methods}

We first demonstrate the practicality of the proposed GoR by integrating it into both logit-based and feature-based KD methods under QAT. We evaluate its effectiveness across diverse tasks, including image classification, OD, and LLM, under standard low-bit settings (e.g., 8-bit). Additionally, we explore more aggressive quantization (e.g., 4-bit) where applicable, excluding scenarios where it leads to severe performance degradation (e.g., OD).

\subsubsection{Classification}

The proposed GoR is a versatile regularization term that integrates seamlessly with existing KD methods. We evaluated its effectiveness by incorporating GoR into WSLD~\citep{zhou2021rethinking}, QFD~\citep{zhu2023quantized}, and SQKD~\citep{zhao24d} for MobileNetV2 and ResNet18 students with a ResNet50 teacher on ImageNet at 8-bit and 4-bit precision. Contrary to SQKD's claim that KD loss alone suffices, adding GoR consistently improves performance (Table~\ref{tab:quantization_results}). For 8-bit quantization, GoR yields modest gains: +0.14\% on MobileNetV2 (71.65\%~\(\rightarrow\)~71.79\%) and +0.11\% on ResNet18 (69.64\%~\(\rightarrow\)~69.75\%). At 4-bit, where quantization challenges are severe, improvements are more pronounced: QAT-KD gains +3.28\% on MobileNetV2 (55.72\%~\(\rightarrow\)~59.01\%) and +0.78\% on ResNet18, while QFD achieves +3.37\% on MobileNetV2. These results highlight GoR's effectiveness, particularly in low-precision scenarios.

\input{classification_table}

\vspace{-0.2cm}
\subsubsection{LLM}

We evaluated our GoR regularizer on LLMs by transferring knowledge from Qwen 2.5 3B (teacher) to Qwen 2.5 0.5B (student) \citep{qwen2025qwen25technicalreport} under 8-bit and 4-bit quantization. We used the FineTome-100k dataset \citep{mlabonne2025finetome100k}, a 100,000-example subset of arcee-ai/The-Tome \citep{arcee-ai2025the-tome}, as our evaluation benchmark. Table~\ref{tab:enhanced_llm_quantization_results} shows performance metrics for Qwen2.5 compression (3B $\rightarrow$ 0.5B) across quantization methods. PTQ exhibited the highest perplexity and lowest downstream metrics, especially at 4-bit. QAT improved upon PTQ at both precisions, while logit distillation (KD) ~\citep{hinton2015distilling, boizard2024towards} further enhanced performance, particularly at 8-bit (perplexity: 5.56). Our proposed GoR Logit Distillation achieved superior results across all metrics, with lowest perplexity (4.89 at 8-bit, 6.27 at 4-bit) and highest BLEU and BertScore, confirming the effectiveness of combining advanced logit distillation with QAT for low-precision quantization.
\input{llm_table}
\subsubsection{Object Detection}
We extended GoR beyond classification to OD by employing YOLOX-Small \citep{ge2021yolox} as the student and YOLOX-Medium \citep{ge2021yolox} as the teacher for QAT-KD, using the COCO dataset \citep{lin2014microsoft} for training and validation. We evaluated state-of-the-art KD for OD methods like CWD \citep{shu2021channelwiseknowledgedistillationdense} and MGD \citep{yang2022maskedgenerativedistillation} alongside baseline KD to benchmark performance. As shown in Table \ref{tab:yolox_detection_results}, GoR consistently improved QAT-KD performance, boosting mean average precision (mAP) at IoU 0.5 from 57.22 to 58.03 (+0.81) for CWD and from 57.68 to 59.2 (+1.52) for MGD. For mAP at IoU 0.5:0.95, GoR achieved improvements from 39.05 to 39.35 (+0.30) for CWD and from 39.25 to 39.48 (+0.23) for MGD. These results highlight GoR's effectiveness in enhancing OD under quantized settings, demonstrating its robustness across different tasks and architectures.
\input{object_table}

\vspace{-0.2cm}
\subsection{Analysis of GoR}
We evaluated learnable regularization against static weighting in QAT-KD (\(\mathcal{L} = (1-\alpha)\mathcal{L}_{\mathrm{task}} + \alpha\mathcal{L}_{\mathrm{KD}}\)) on ImageNet with MobileNetV2 (8-bit). As shown in Table~\ref{tab:gorvsnonlearnable}, static weights achieved Top-1 accuracies of 71.46\% (\(\alpha=1.0\)), 71.52\% (\(\alpha=0.2\)), and 71.62\% (\(\alpha=0.5\)), while our GoR framework with learnable \(\alpha_{\mathrm{task}}\) and \(\alpha_{\mathrm{KD}}\) achieved 71.79\%. Furthermore, Table~\ref{tab:performance_reweight} shows that GoR outperforms existing reweighting methods, such as dynamic knowledge distillation (DKD) ~\citep{li2021dynamic}, demonstrating its adaptability across various models, datasets, and quantization schemes without requiring manual tuning.



\begin{figure}[!htb]

  \begin{minipage}[t]{0.27\linewidth}
      \centering
      \includegraphics[width=\linewidth, height=2.6cm]{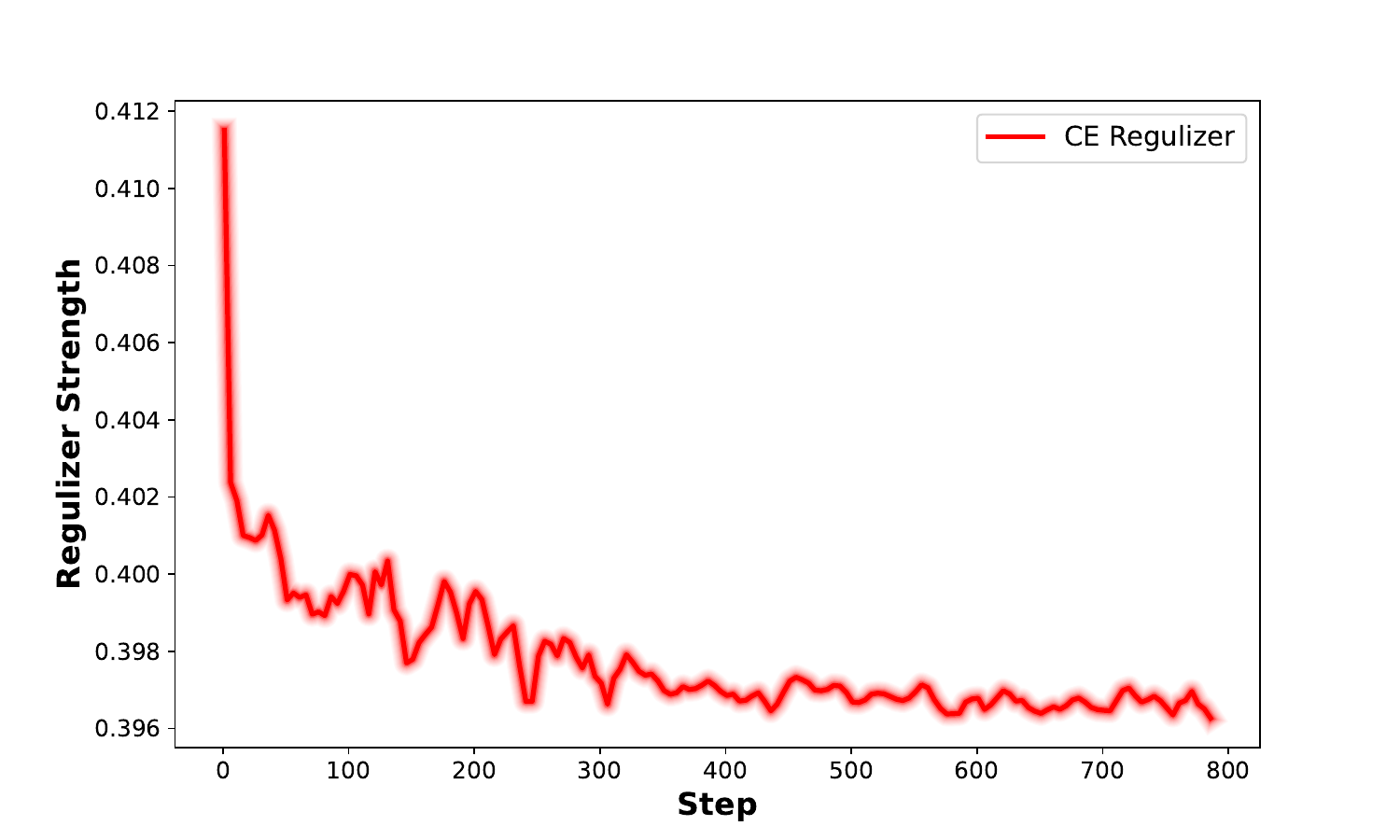}
      \small{(a)}
  \end{minipage}
  \hfill
  \hspace{-.6cm}
  \begin{minipage}[t]{0.27\linewidth}
      \centering
      \includegraphics[width=\linewidth, height=2.6cm]{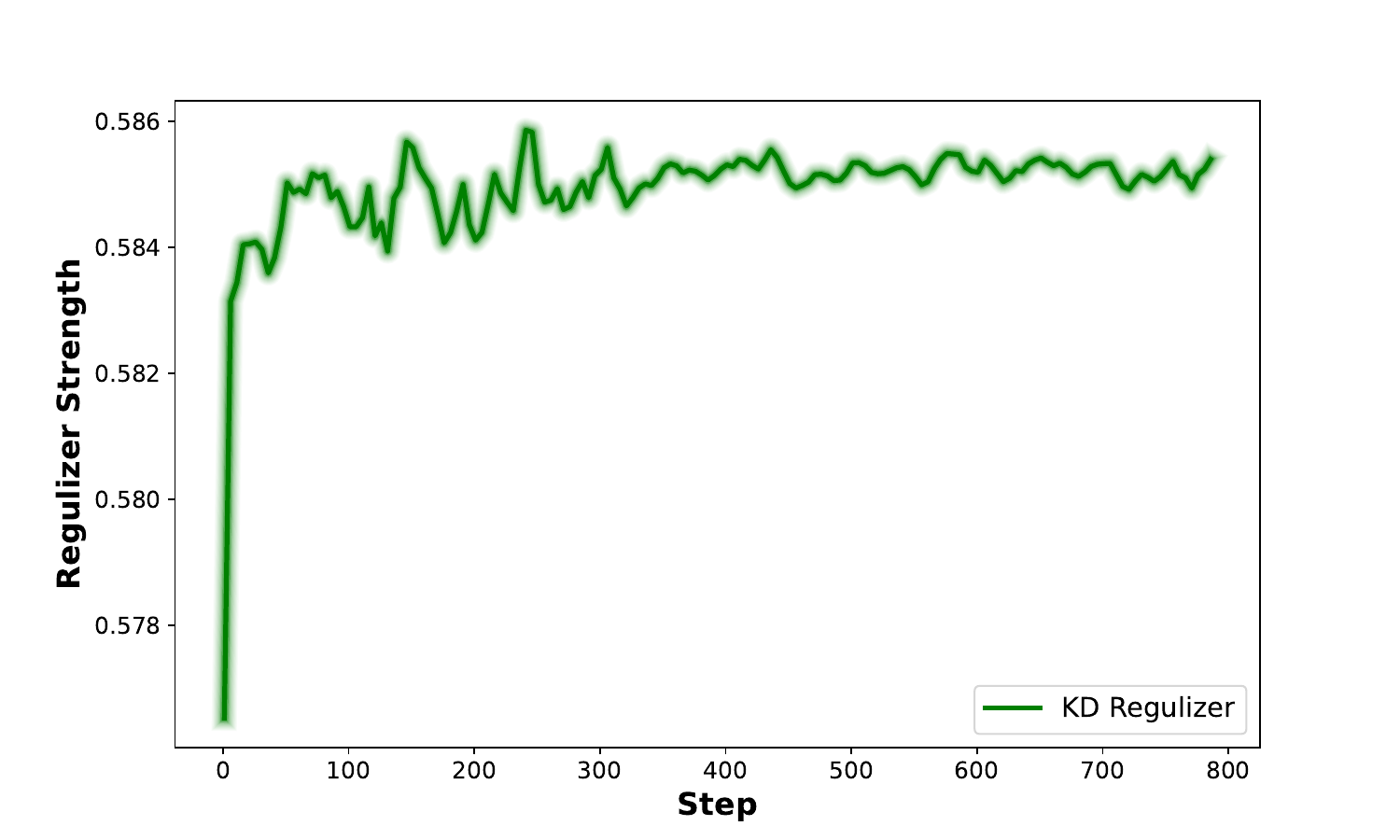}
      \small{(b)}
  \end{minipage}
  \hfill
  \hspace{-.6cm}
  \centering
    \begin{minipage}[t]{0.27\linewidth}
      \centering
      \includegraphics[width=\linewidth, height=2.6cm]{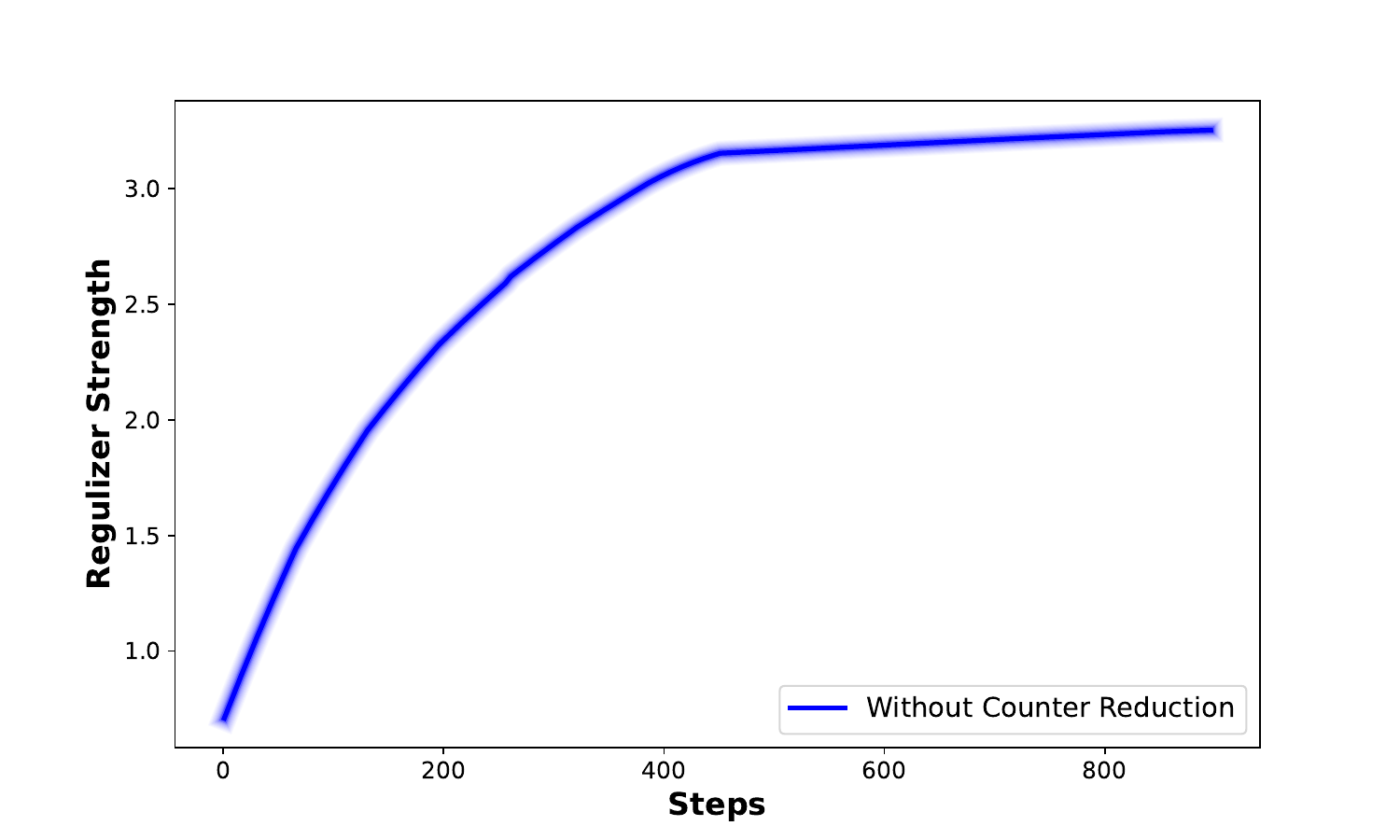}
      \small{(c)}
  \end{minipage}
  \hfill
  \hspace{-.35cm}
  \begin{minipage}[t]{0.23\linewidth}
      \centering
      \includegraphics[width=\linewidth, height=2.36cm]{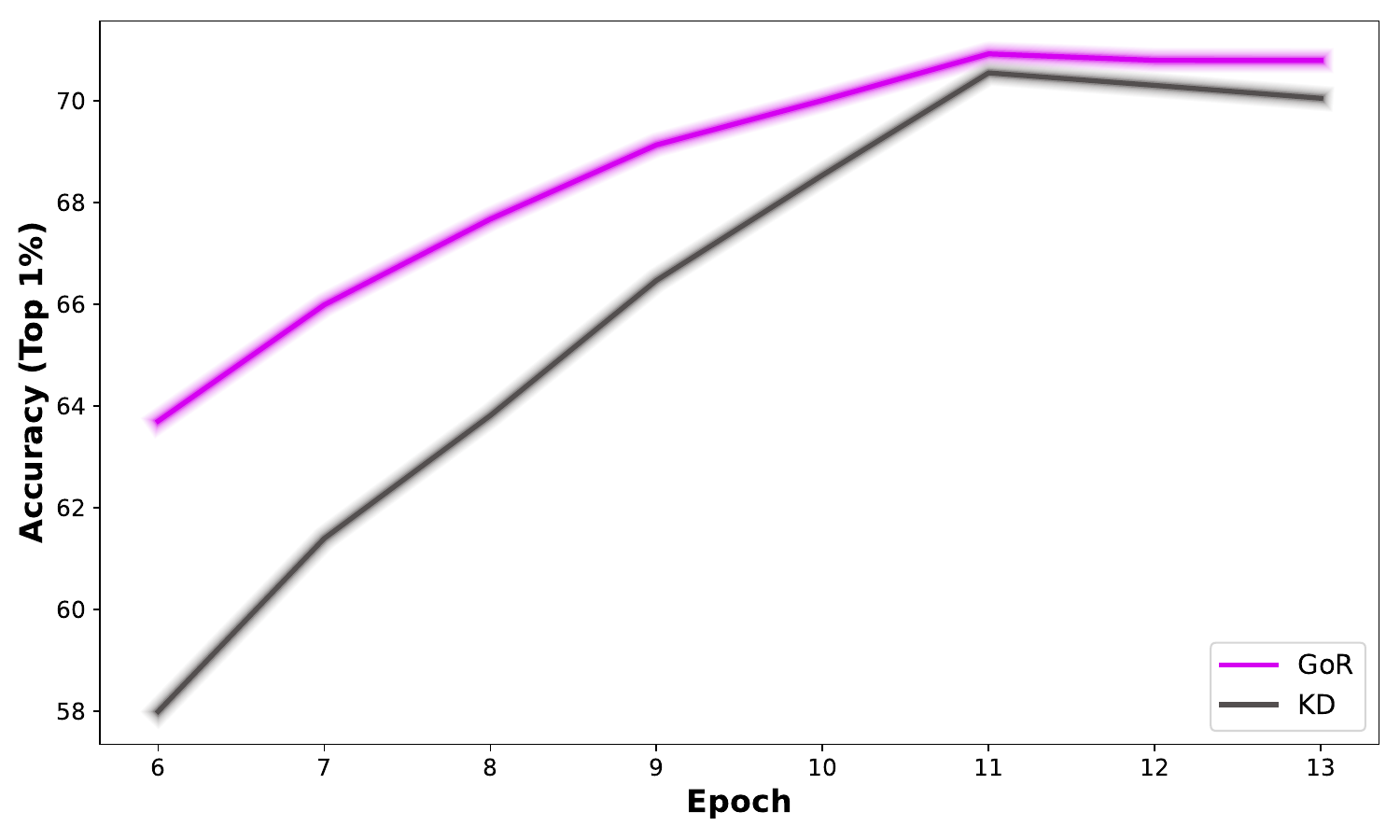}
      \small{(d)}
  \end{minipage}

  \caption{Effect of GoR on QAT-KD training stability for MobileNet-V2: (a-b) Balanced training with counter-reduction, (c) Instability without counter-reduction, and (d) convergence acceleration via dynamic regularization.}
  \label{fig:analysis}
  \vspace{-0.5cm}
\end{figure}

\begin{table}[!htb]
\centering
\begin{minipage}[t]{0.4\textwidth}
\centering
\caption{Comparison of GoR with static regularizers on QAT-KD.}
\scalebox{0.65}{
\begin{tabular}{ccc}
\hline
\textbf{Method} & \textbf{Weight ($\alpha$)} & \textbf{Top-1 (\%)} \\
\hline
\multirow{3}{*}{Static weighting} & 1.0 & 71.46 \\
& 0.2 & 71.52 \\
& 0.5 & 71.62 \\
\hline
GoR (Ours) & Learnable & \textbf{71.79} \\
\hline
\end{tabular}}
\label{tab:gorvsnonlearnable}
\end{minipage}
\hfill
\begin{minipage}[t]{0.56\textwidth}
\centering
\caption{Performance comparison with reweighting method on ImageNet and CIFAR-100 with MobileNet-v2.}
\scalebox{0.6}{
\begin{tabular}{ccc}
\hline
\textbf{Method} & \textbf{ImageNet} & \textbf{CIFAR-100} \\
\hline
QAT + KD & 71.65 & 76.71 \\
QAT + Reweighting (DKD) \citep{li2021dynamic} & 71.52 (-0.13) & 76.71 (+0.0) \\
QAT + KD + GoR & \textbf{71.79} \color{blue}(+0.14) & \textbf{76.90} \color{blue}(+0.19) \\
\hline
\end{tabular}}
\label{tab:performance_reweight}
\end{minipage}
\end{table}
Moreover, Figure~\ref{fig:analysis} visualizes the training dynamics of the QAT-KD for MobileNet-V2. Figures~\ref{fig:analysis}(a) and \ref{fig:analysis}(b) illustrate the CE and KD losses with counter-reduction, respectively, showing that our approach successfully stabilizes training by dynamically balancing the interactions between loss components. Conversely, Figure~\ref{fig:analysis}(c) highlights the scenario without counter-reduction, where a single learnable regularizer keeps getting larger before getting clipped, leading to excessive amplification of quantization-induced distortions, hindering convergence and reducing overall accuracy. Figure~\ref{fig:analysis}(d) highlights the QAT-KD method along with the integration of dynamic regularization through GoR, which accelerates faster convergence.

\begin{figure}[!htb]
  \centering

   \begin{minipage}[t]{0.27\linewidth}
    \centering
    \vspace{-2.23cm} 
    \includegraphics[width=\linewidth, height=4.65cm]{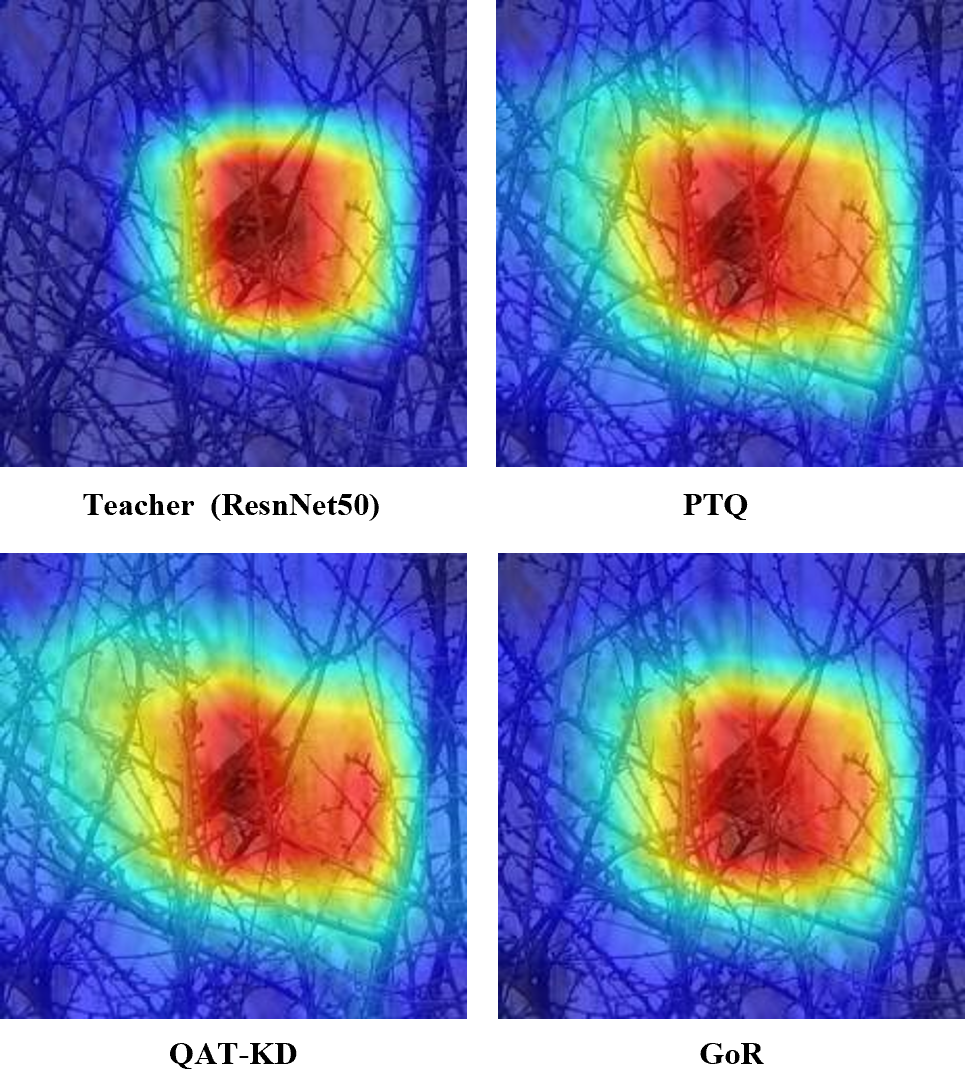}\\
    \small{(a)}
  \end{minipage}
  \begin{minipage}[t]{0.51\linewidth}
    \centering
    \includegraphics[width=0.47\linewidth]{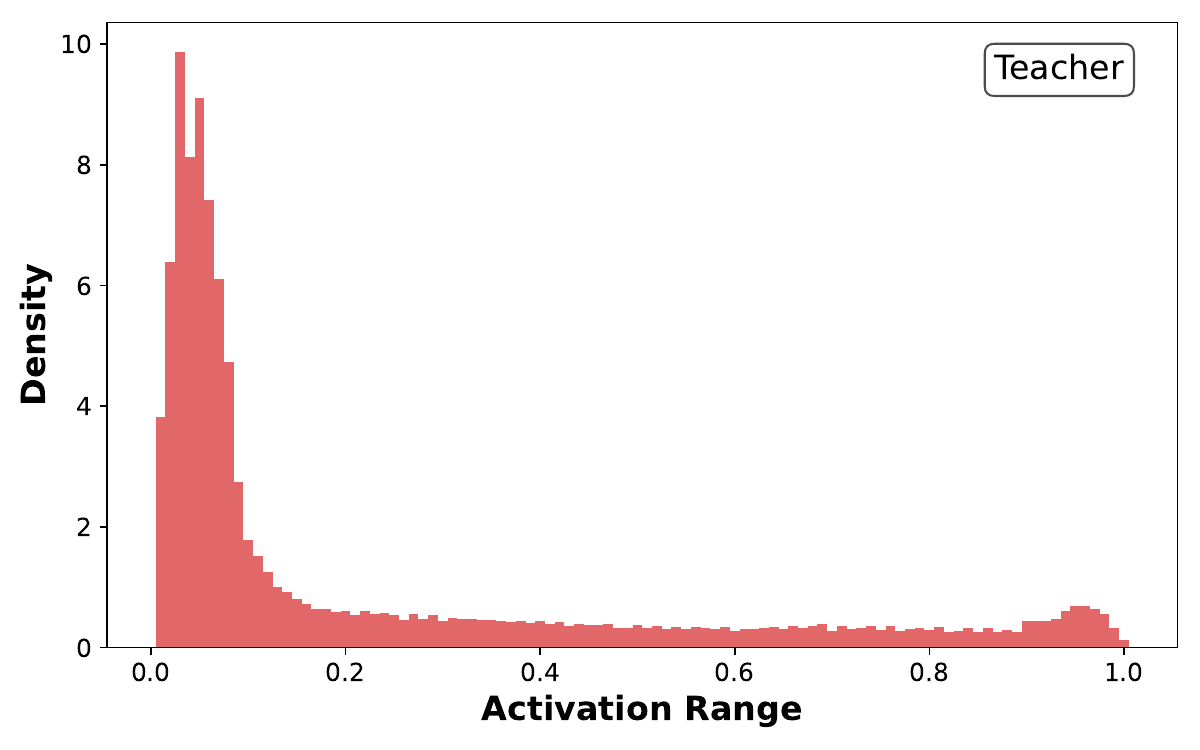}%
    \includegraphics[width=0.47\linewidth]{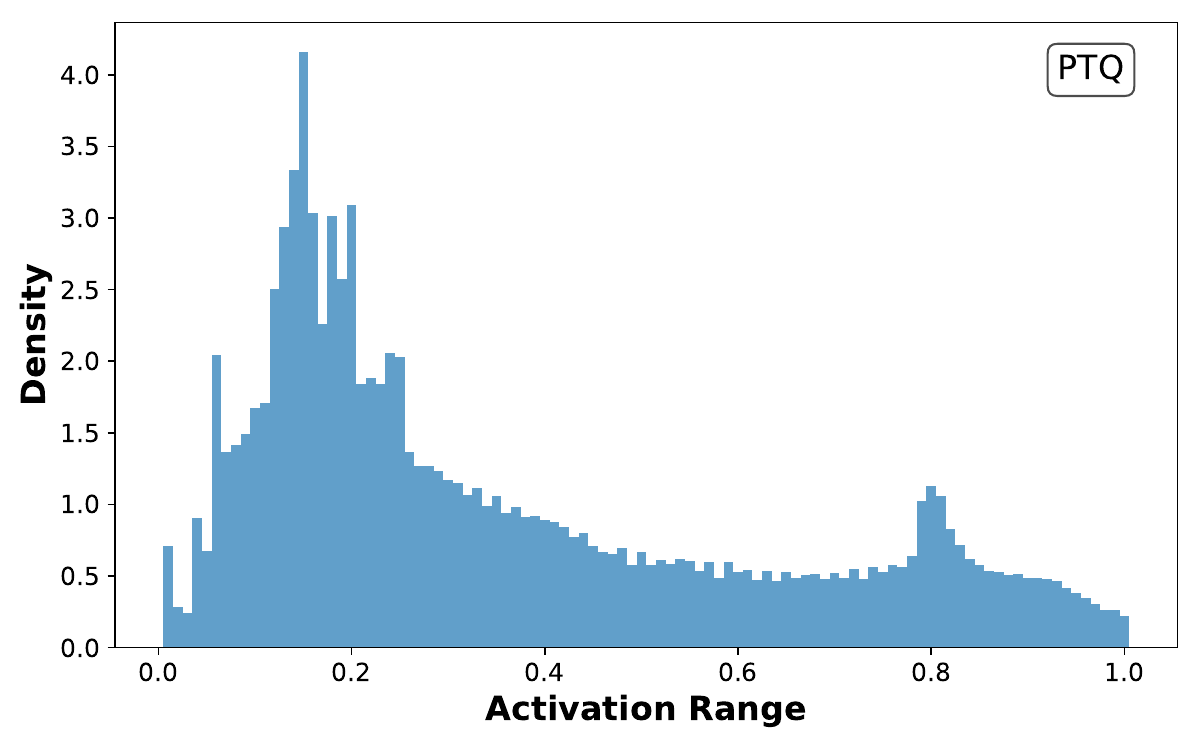}\\[0.3em]
    \includegraphics[width=0.47\linewidth]{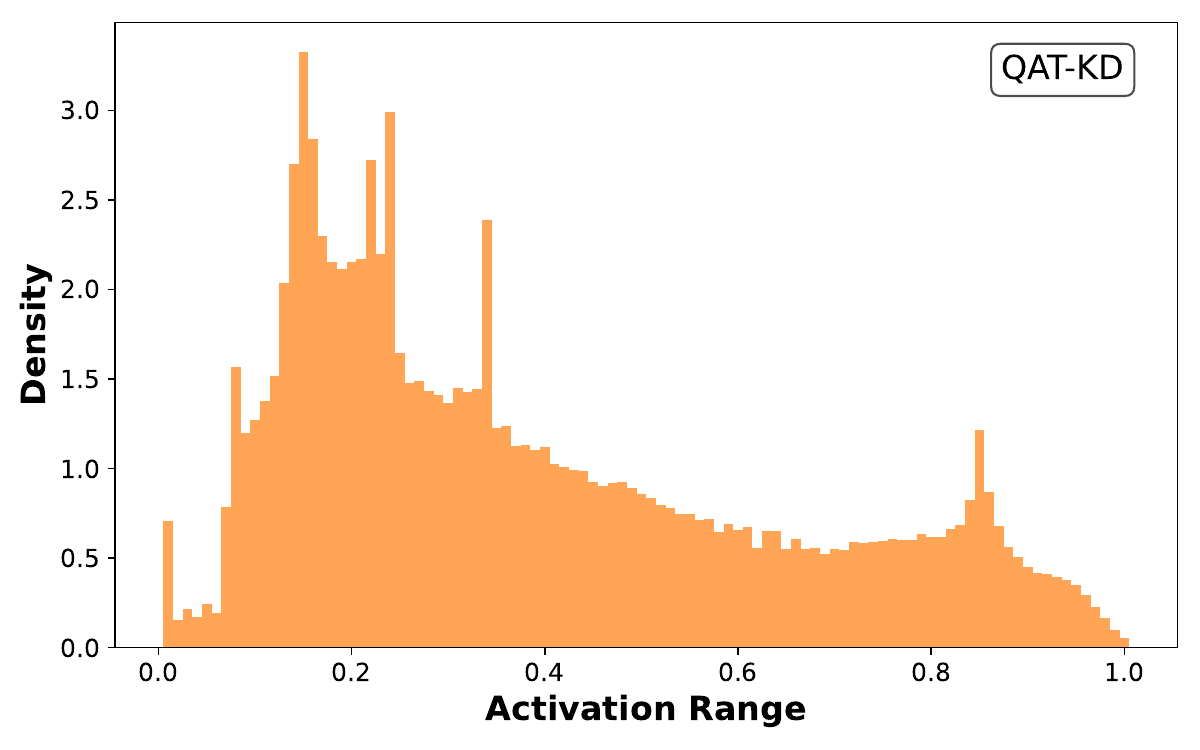}%
    \includegraphics[width=0.47\linewidth]{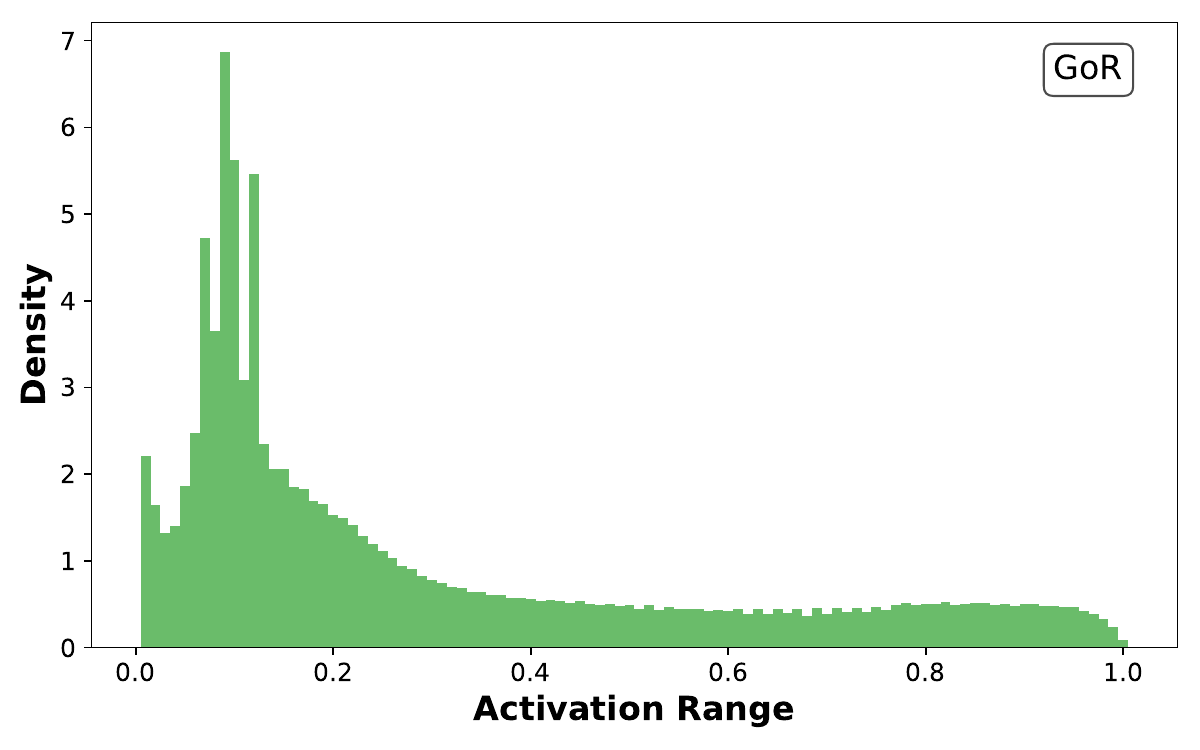}\\
    \small{(b)}
  \end{minipage}%
  \vspace{-.15cm}
  \hspace{-0.01\linewidth}%
  \begin{minipage}[t]{0.2\linewidth}
    \centering
    \includegraphics[width=\linewidth]{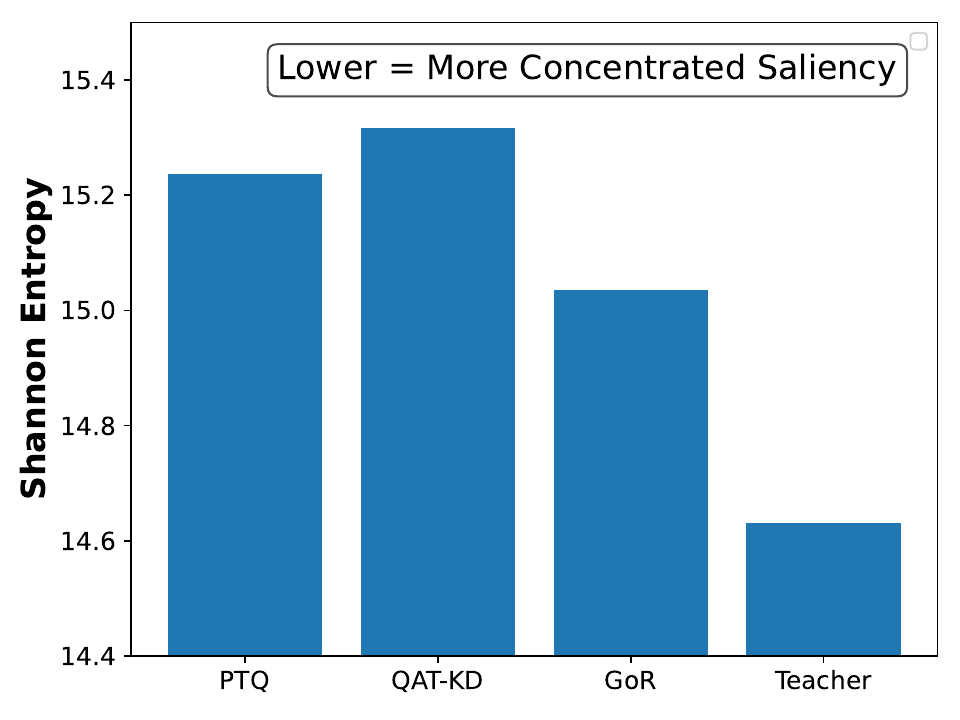}\\[0.3em]
    \includegraphics[width=\linewidth]{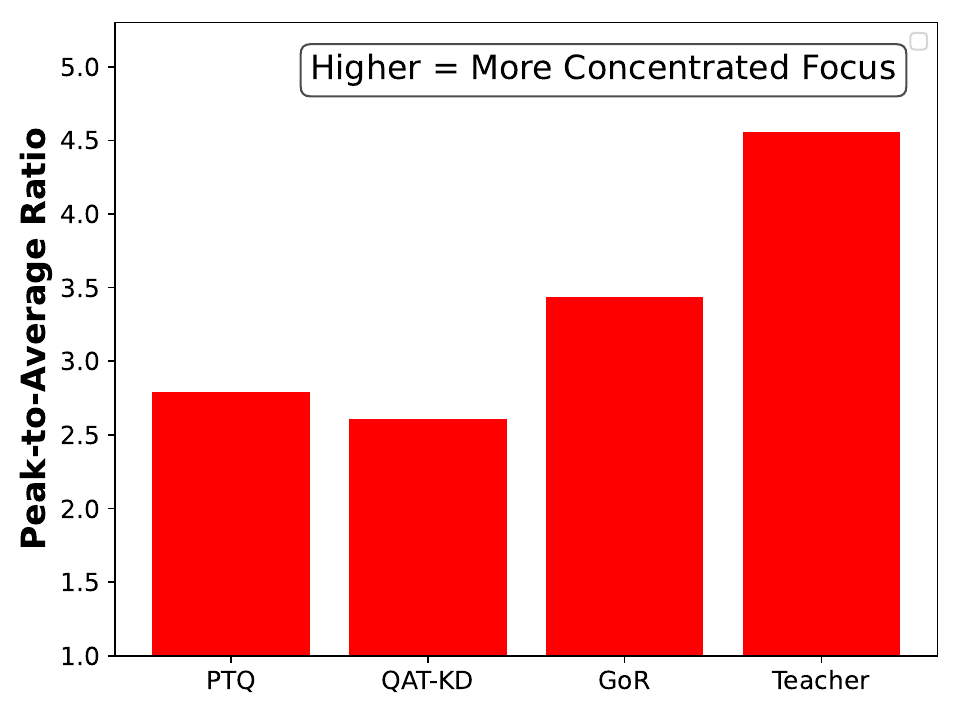}\\
    \small{(c)}
  \end{minipage}%
  \hspace{0.01\linewidth}%
 
  \caption{GoR-QAT-KD outperforms PTQ and QAT-KD in attention preservation: (a) Grad-CAM visualizations, (b) activation distributions, and (c) attention metrics.}
  \label{fig:feat_vis}
  \vspace{-.5cm}
\end{figure}

Figure~\ref{fig:feat_vis} compares attention patterns across teacher, PTQ, QAT-KD, and GoR-QAT-KD models. Figure~\ref{fig:feat_vis}(a) shows GradCAM~\citep{selvaraju2017grad} activation maps where GoR-QAT-KD closely matches the teacher's localization patterns. Figure~\ref{fig:feat_vis}(b) presents activation intensity histograms, with GoR-QAT-KD exhibiting similar distribution to the teacher. Figure~\ref{fig:feat_vis}(c) quantifies attention quality using peak-to-average ratio (measuring attention concentration) and Shannon entropy (capturing attention diversity). These complementary metrics illustrate that the SQM optimized with the proposed method preserves the teacher's balanced focus on salient regions. GoR-QAT-KD achieves values closest to those of the teacher for both metrics, while PTQ and QAT-KD exhibit larger deviations, confirming superior attention preservation.




\subsection{EKD with GoR}
We extend our evaluation of GoR beyond QAT-KD by exploring its integration with EKD under QAT, targeting edge-efficient models like MobileNetV2 that typically lack scalable variants and homogeneous teacher options. To address this, we leverage heterogeneous teacher ensembles (e.g., RegNet, ConvNeXt, Swin-T) within the EKD-GoR framework. As shown in Table~\ref{tab:single_multimodel}, our method achieves 71.99\% Top-1 accuracy, outperforming both single-teacher KD and homogeneous EKD setups—and even surpassing the full-precision baseline (71.87\%). 

We also compare the proposed EKD-GoR with existing multi-teacher methods under QAT. As shown in Table~\ref{tab:multi_teacher_comparison}, EKD-GoR achieves the highest accuracy at both 8-bit (71.99\%) and 4-bit (59.87\%) precision, significantly outperforming recent approaches such as  AMT~\citep{liu2020adaptive},  CMT~\citep{wang2021collaborative}, and  CAMKD~\citep{zhang2022confidence}. The substantial gain at 4 bits highlights the robustness and practicality of EKD-GoR for real-world low-bit deployments.
\input{multi_teacher_table}

\subsection{Real-world Edge Performance}

We evaluated the proposed QAT-KD-GoR framework by deploying the optimized SQM on real-edge hardware, such as the Jetson Orin. To generalize beyond the MobileNet–ImageNet setup, we extended our study to include commonly used lightweight architectures such as VGG, ResNet18, and RegNetX-400MF~\citep{radosavovic2020designing}, as well as transformer-based models like Swin-T~\citep{liu2021swin}. We further examined its applicability to full-precision training on smaller datasets like CIFAR-100, using transfer learning with ImageNet-pretrained weights while maintaining consistency. Table~\ref{tab:final_results_fps} reports accuracy and FPS across ImageNet and CIFAR-100 under various Jetson Orin power settings (MaxN, 15W, 30W, 50W). The QAT-KD-GoR framework consistently outperformed both baseline and standard QAT+KD. For example, ResNet50 as a teacher yielded a 0.47\% gain for ResNet18 on ImageNet and improved FPS; RegNet improved accuracy by 0.24\% with better performance at lower power. Swin-T achieved a 0.26\% gain on ImageNet, along with notable FPS improvements, especially at 15W. On CIFAR-100, Swin-T showed the largest gain of 1.67\%. These results demonstrate the robustness and deployment readiness of the QAT-KD-GoR framework across architectures, datasets, and power-constrained environments. Notably, the proposed method can compress SQMs to achieve up to a 242\% increase in inference speed compared to the base model, without any loss in accuracy—and in some cases, even achieving better accuracy than the full-precision baseline.

\input{generalization}

\subsection{Discussion}
We demonstrate the practicality of the GoR-driven QAT-KD framework for optimizing SQM in real-world applications. Our results show that a learnable regularizer with just two parameters can significantly enhance existing KD methods under QAT. Additionally, leveraging heterogeneous teachers via EKD-GoR narrows the performance gap with full-precision models and can even surpass baseline models under optimal conditions. QAT+KD+GoR is a versatile strategy that could be applicable to various vision tasks, including image-to-image translation, segmentation, detection, and LLMs. We plan to explore its broader applications in future work.

\section{Conclusion}

In this study, we propose a novel learnable regularization strategy to optimize SQMs. We demonstrate that only two learnable parameters can effectively address the well-known QAT-KD loss balancing issues. This results in significantly reduced quantization error, enabling efficient knowledge transfer from large models to SQMs. This strategy boosts the performance of existing methods across various AI tasks and domains under aggressive quantization. Furthermore, by incorporating EKD with multiple heterogeneous teacher models, our framework further refines the student’s distribution and can even surpass full-precision models under optimal conditions. We plan to explore the broader applicability of the proposed EKD-GoR framework to diverse vision and AI tasks in future work.


\bibliography{acml25}






\end{document}

%% file: classification_table.tex
\begin{table}[htbp]
\centering
\caption{Comparison of QAT-KD methods on classification with and without GoR for MobileNetV2 and ResNet18.}
\label{tab:quantization_results}
\resizebox{\textwidth}{!}{
\begin{tabular}{lcccccccc}
\toprule
\multirow{3}{*}{\textbf{Method}} & \multicolumn{4}{c}{\textbf{ResNet50 $\xrightarrow{}$ MobileNetV2}} & \multicolumn{4}{c}{\textbf{ResNet50 $\xrightarrow{}$ ResNet18}} \\ 
\cmidrule(lr){2-5} \cmidrule(lr){6-9}
 & \multicolumn{2}{c}{$\mathbf{8/8}$} & \multicolumn{2}{c}{$\mathbf{4/4}$} & \multicolumn{2}{c}{$\mathbf{8/8}$} & \multicolumn{2}{c}{$\mathbf{4/4}$} \\ 
\cmidrule(lr){2-3} \cmidrule(lr){4-5} \cmidrule(lr){6-7} \cmidrule(lr){8-9}
 & w/o GoR & GoR & w/o GoR & GoR & w/o GoR & GoR & w/o GoR & GoR \\ 
\midrule
PTQ & 71.08 & -- & 0.70 & -- & 69.39 & -- & 0.35 & -- \\[2pt]
QAT & 71.36 & -- & 43.82 & -- & 69.56 & -- & 60.30 & -- \\[2pt]
QAT + KD ~\citep{hinton2015distilling}
& 71.65 & \textbf{71.79} {\color{blue}\small{(+0.14)}} 
& 55.72 & \textbf{59.01} {\color{blue}\small{(+3.28)}} 
& 69.64 & \textbf{69.75} {\color{blue}\small{(+0.11)}} 
& 60.80 & \textbf{61.57} {\color{blue}\small{(+0.78)}} \\

WSLD \cite{zhou2021rethinking} 
& 71.60 & \textbf{71.71} {\color{blue}\small{(+0.11)}} 
& 47.91 & \textbf{48.35} {\color{blue}\small{(+0.45)}} 
& 69.65 & \textbf{69.71} {\color{blue}\small{(+0.06)}} 
& 61.91 & \textbf{62.09} {\color{blue}\small{(+0.18)}} \\[2pt]

QFD \cite{zhu2023quantized} 
& 71.73 & \textbf{71.76} {\color{blue}\small{(+0.03)}} 
& 56.00 & \textbf{59.38} {\color{blue}\small{(+3.37)}} 
& 69.73 & \textbf{69.86} {\color{blue}\small{(+0.13)}} 
& 60.95 & \textbf{61.18} {\color{blue}\small{(+0.23)}} \\ 

SQKD \cite{zhao24d} 
& 71.43 & \textbf{71.54} {\color{blue}\small{(+0.11)}} 
& 47.13 & \textbf{48.84} {\color{blue}\small{(+1.71)}} 
& 69.56 & \textbf{69.75} {\color{blue}\small{(+0.21)}} 
& 60.94 & \textbf{61.16} {\color{blue}\small{(+0.22)}} \\ 
\bottomrule
\end{tabular}}
\end{table}

%% file: llm_table.tex
\begin{table}[htbp]
\centering
\caption{Enhanced quantization results for LLM (Qwen2.5 3B $\rightarrow$ Qwen2.5 0.5B) at 8-bit and 4-bit precision, showing improvements.}
\label{tab:enhanced_llm_quantization_results}
\resizebox{\textwidth}{!}{%
\begin{tabular}{lccccc|ccccc}
\toprule
\multirow{2}{*}{\textbf{Method}} & \multicolumn{5}{c|}{\textbf{8/8}} & \multicolumn{5}{c}{\textbf{4/4}} \\ 
\cmidrule(lr){2-6}\cmidrule(lr){7-11}
& Perplexity $\downarrow$ & MMLU $\uparrow$ & Hellaswag $\uparrow$ & BLEU $\uparrow$ & BertScore $\uparrow$ & Perplexity $\downarrow$ & MMLU $\uparrow$ & Hellaswag $\uparrow$ & BLEU $\uparrow$ & BertScore $\uparrow$ \\ 
\midrule
PTQ                  
& 7.74 & 23.15 & 30.15 & 7.94 & 77.6 
& 9.56 & 22.87 & 28.70 & 7.53 & 75.01 \\[2pt]

QAT                  
& 6.57 & 26.46 & 31.33 & 8.56 & 83.17 
& 6.85 & 22.95 & 29.74 & 8.02 & 77.84 \\[2pt]

QAT + KD ~\citep{boizard2024towards}
& 5.55 & 27.87 & \textbf{33.14} & 9.96 & 83.50 
& 8.50 & 23.41 & 30.81 & 8.31 & 79.12 \\[2pt]

\textbf{QAT + KD + GoR (Ours)} 
& \textbf{4.89} & \textbf{28.38} & 32.81 & \textbf{12.73} & \textbf{85.96}
& \textbf{6.27} & \textbf{23.94} & \textbf{31.52} & \textbf{8.84} & \textbf{83.15} \\[2pt]

& {\color{blue}(-0.66)} & {\color{blue}(+0.51)} & {\color{blue}(-0.33)} & {\color{blue}(+2.77)} & {\color{blue}(+2.46)}
& {\color{blue}(-0.58)} & {\color{blue}(+0.53)} & {\color{blue}(+0.71)} & {\color{blue}(+0.53)} & {\color{blue}(+4.03)} \\
\bottomrule
\end{tabular}}
\end{table}

%% file: object_table.tex
\begin{table}[htbp]
\centering
\caption{Quantization performance comparison for YOLOX-M $\rightarrow$ YOLOX-S detection with and without GoR.}
\label{tab:yolox_detection_results}
\resizebox{0.56\textwidth}{!}{%
\begin{tabular}{lcccc}
\toprule
\multirow{2}{*}{\textbf{Method}} & \multicolumn{2}{c}{\textbf{w/o GoR}} & \multicolumn{2}{c}{\textbf{GoR}} \\ 
\cmidrule(lr){2-3}\cmidrule(lr){4-5}
 & \textbf{AP50} & \textbf{AP50:95} & \textbf{AP50} & \textbf{AP50:95} \\ 
\midrule
PTQ            & 56.81 & 38.75 & -- & -- \\
QAT            & 57.23 & 39.05 & -- & -- \\
CWD \citep{shu2021channelwiseknowledgedistillationdense}   & 57.22 & 39.05 & \textbf{58.03 {\color{blue}(+0.81)}} & \textbf{39.35 {\color{blue}(+0.30)}} \\
MGD \citep{yang2022maskedgenerativedistillation}   & 57.68 & 39.25 & \textbf{59.20 {\color{blue}(+1.52)}} & \textbf{39.48 {\color{blue}(+0.23)}} \\ 
\bottomrule
\end{tabular}}
\end{table}

%% file: multi_teacher_table.tex
\begin{table}[htbp]
\centering
\begin{minipage}{0.65\textwidth}
\centering
\caption{Single vs multi-teacher performance comparison.}
\label{tab:single_multimodel}
\scalebox{0.57}{%
\begin{tabular}{lccccc}
\toprule
\multirow{2}{*}{\textbf{Method}} & \multirow{2}{*}{\textbf{Teacher Type}} & \multicolumn{3}{c}{\textbf{Teacher}} & \textbf{Student} \\ 
\cmidrule(lr){3-5}\cmidrule(lr){6-6}
 & & \textbf{Model} & \textbf{Params (M)} & \textbf{Top-1} & \textbf{Top-1} \\ 
\midrule
Baseline                & N/A    & --         & --     & --       & 71.87           \\
PTQ                     & N/A    & --         & --     & --       & 71.08 \\
QAT                     & N/A    & --         & --     & --       & 71.36 \\
\midrule
QAT + KD                      & Single & ResNet50   & 25.6   & 76.13    & 71.66 {\color{blue}(+0.36)}   \\
QAT + KD                      & Single & RegNet     & 5.5    & 72.83    & 71.51 {\color{blue}(+0.15)}   \\
QAT + KD                      & Single & ConvNeXt   & 28.6   & 82.52    & 71.43 {\color{blue}(+0.08)}   \\
QAT + KD                      & Single & Swin-T     & 28.3   & 81.47    & 71.28 {\color{blue}(-0.08)}   \\
\midrule
\multirow{2}{*}{\shortstack{QAT + EKD + GoR \\ (Homogeneous)}}& Multi  & ResNet50   & 25.6   & 76.13    & \multirow{2}{*}{71.82 {\color{blue}(+0.46)}} \\
                        & Multi  & ResNet101  & 44.50   & 77.37    &                 \\
\midrule
\multirow{3}{*}{\shortstack{QAT + EKD + GoR \\ (Proposed)}}  & Multi  & RegNet     & 5.5    & 72.83    & \multirow{3}{*}{71.99 {\color{blue}(+0.63)}} \\
                        & Multi  & ConvNeXt   & 28.6   & 82.52    &                 \\
                        & Multi  & Swin-T     & 28.3   & 81.47    &                 \\
\bottomrule
\end{tabular}}
\end{minipage}\hfill
\begin{minipage}{0.34\textwidth}
\centering
\caption{Comparison between multi-teacher KD and the proposed EKD-GoR.}
\label{tab:multi_teacher_comparison}
\scalebox{.65}{\begin{tabular}{lcc}
\toprule
\textbf{Method} & \textbf{8/8} & \textbf{4/4} \\ 
\midrule
AMT \citep{liu2020adaptive}                    & 71.58 & 53.17 \\
CMT \citep{wang2021collaborative}                    & 71.71 & 55.75 \\
CAMKD \citep{zhang2022confidence}                  & 71.33 & 55.46 \\
\textbf{EKD + GoR (Ours)} & \textbf{71.99} & \textbf{59.87} \\ 
\bottomrule
\end{tabular}}
\end{minipage}
\end{table}

%% file: generalization.tex
\begin{table}[htbp]
\centering
\caption{Quantization accuracy (Top-1 \%) with FPS across ImageNet and CIFAR100 datasets. Accuracy gains over baseline/QAT+KD shown in parentheses.}
\label{tab:final_results_fps}
\resizebox{\textwidth}{!}{%
\begin{tabular}{lcccccccccccccccc}
\toprule
\multirow{2}{*}{\textbf{Model}} & \multicolumn{4}{c}{\textbf{ImageNet }} & \multicolumn{4}{c}{\textbf{CIFAR100}} & \multicolumn{8}{c}{\textbf{Speed on Jetson (FPS)}} \\ 
\cmidrule(lr){2-5} \cmidrule(lr){6-9} \cmidrule(lr){10-17}
 & Base & KD + GoR (FP32) & QAT+KD & Our (INT8) & Base & KD + GoR (FP32) & QAT+KD & Our (INT8) & MaxN FP32 & MaxN INT8 & 15W FP32 & 15W INT8 & 30W FP32 & 30W INT8 & 50W FP32 & 50W INT8 \\ 
\midrule
Resnet18
& 69.76 & \textbf{70.23} {\color{blue}(+0.47)} 
& 69.64 & \textbf{69.71} {\color{blue}(+0.11)} 
& 80.02 & \textbf{80.19} {\color{blue}(+0.17)} 
& 79.89 & \textbf{80.01} {\color{blue}(+0.12)} 
& 1087 & 2447 & 167 & 623 & 381 & 1213 & 736 & 1856 \\[2pt]

VGG 
& 71.59 & \textbf{71.65} {\color{blue}(+0.06)} 
& 71.52 & \textbf{71.64} {\color{blue}(+0.12)} 
& 78.95 & \textbf{79.48} {\color{blue}(+0.53)} 
& 78.93 & \textbf{79.22} {\color{blue}(+0.29)} 
& 166 & 568 & 30 & 121 & 60 & 229 & 121 & 414 \\[2pt]

MobileNet-v2 
& 71.87 & \textbf{72.16} {\color{blue}(+0.29)} 
& 71.65 & \textbf{71.79} {\color{blue}(+0.14)} 
& 76.83 & \textbf{77.07} {\color{blue}(+0.24)} 
& 76.71 & \textbf{76.90} {\color{blue}(+0.19)} 
& 948 & 1103 & 182 & 295 & 338 & 461 & 578 & 717 \\[2pt]

RegNetX-400MF 
& 72.83 & \textbf{73.07} {\color{blue}(+0.24)} 
& 72.79 & \textbf{72.97} {\color{blue}(+0.18)} 
& 81.73 & \textbf{81.90} {\color{blue}(+0.17)} 
& 81.57 & \textbf{81.70} {\color{blue}(+0.13)} 
& 1169 & 1805 & 196 & 491 & 456 & 823 & 733 & 1151 \\[2pt]

Swin-T 
& 81.47 & \textbf{81.73} {\color{blue}(+0.26)} 
& 81.46 & \textbf{81.59} {\color{blue}(+0.13)} 
& 80.32 & \textbf{81.99} {\color{blue}(+1.67)} 
& 80.29 & \textbf{80.50} {\color{blue}(+0.21)} 
& 117 & 134 & 20 & 25 & 37 & 45 & 73 & 86 \\ 
\bottomrule
\end{tabular}}
\end{table}